\documentclass[10pt,twocolumn,letterpaper]{article}

\usepackage[pagenumbers]{cvpr} 

\usepackage{graphicx}
\usepackage{amsmath}
\usepackage{amssymb}
\usepackage{booktabs}

\usepackage{times}
\usepackage{epsfig}
\usepackage{float}
\usepackage{algorithm}
\usepackage{algpseudocode}

\usepackage{multirow}

\usepackage{colortbl}
\usepackage{bm}

\usepackage[pagebackref,breaklinks,colorlinks]{hyperref}

\usepackage{adjustbox}

\newcommand*{\affaddr}[1]{#1} 
\newcommand*{\affmark}[1][*]{\textsuperscript{#1}}

\usepackage[capitalize]{cleveref}
\crefname{section}{Sec.}{Secs.}
\Crefname{section}{Section}{Sections}
\Crefname{table}{Table}{Tables}
\crefname{table}{Tab.}{Tabs.}

\def\cvprPaperID{*****} 
\def\confName{CVPR}
\def\confYear{2022}

\title{Topic Detection in Continuous Sign Language Videos}

\vspace{2cm}

\author{%
Álvaro Budria\affmark[1] \quad Laia Tarrés\affmark[1] \quad Gerard I. Gállego\affmark[1] \vspace{0.15cm}\\
Francesc Moreno-Noguer\affmark[2] \quad Jordi Torres\affmark[1,3] \quad Xavier Giró-i-Nieto\affmark[1,2] \vspace{0.1cm}\\
\affaddr{\affmark[1]{\footnotesize \textit{Universitat Politècnica de Catalunya \quad}}} 
\affaddr{\affmark[2]{\footnotesize \textit{Institut de Robòtica i Informàtica Industrial, CSIC-UPC \quad}}} 
\affaddr{\affmark[3]{\footnotesize \textit{Barcelona Supercomputing Center}}}
}

\begin{document}

\maketitle


\begin{abstract}
   \vspace{-0.2cm}
   Significant progress has been made recently on challenging tasks in automatic sign language understanding, such as sign language recognition, translation and production. However, these works have focused on datasets with relatively few samples, short recordings and limited vocabulary and signing space. 
   In this work, we introduce the novel task of sign language topic detection. We base our experiments on How2Sign \cite{Duarte_CVPR2021}, a large-scale video dataset spanning multiple semantic domains. We provide strong baselines for the task of topic detection,
   and present a comparison between different visual features commonly used in the domain of sign language.
   \vspace{-0.35cm}
\end{abstract}


\section{Introduction}
\label{sec:intro}

Sign languages are the native languages and primary means of communication for millions of Deaf and hard-of-
hearing people worldwide. Sign languages utilize multiple complementary channels to convey information, including manual features, such as shape, movement and pose, as well as non-manual features, such as facial expressions and movement of head, shoulders and torso.

Tasks of diverse complexity have been addressed in the literature: from the simpler sign language recognition \cite{SLRdistillation, SkeletonSLR, RGB_SLR, SLR_coArt, SLR_VizAlign, SLR_CrossModAlign, SLR_stochastic} over isolated signs, to the much more challenging ones of sign language translation \cite{Phoenix2, SLT, SLT_MultiChannel, SLT_Korean} and production \cite{SLP_Photo, SLP_GraphSAtt, SLP_Mixed, SLP_ProgTransf, SLT_EvSignNow}. While some methods for translation and production have shown very good results on smaller datasets \cite{AUTSL, WLASL, SLT_Korean, Phoenix, Phoenix2}, they have not been proven to produce satisfactory results yet on larger ones containing a wider signing space with longer video sequences.

\begin{figure}[t]
    \begin{center}
        \includegraphics[width=0.95\linewidth]{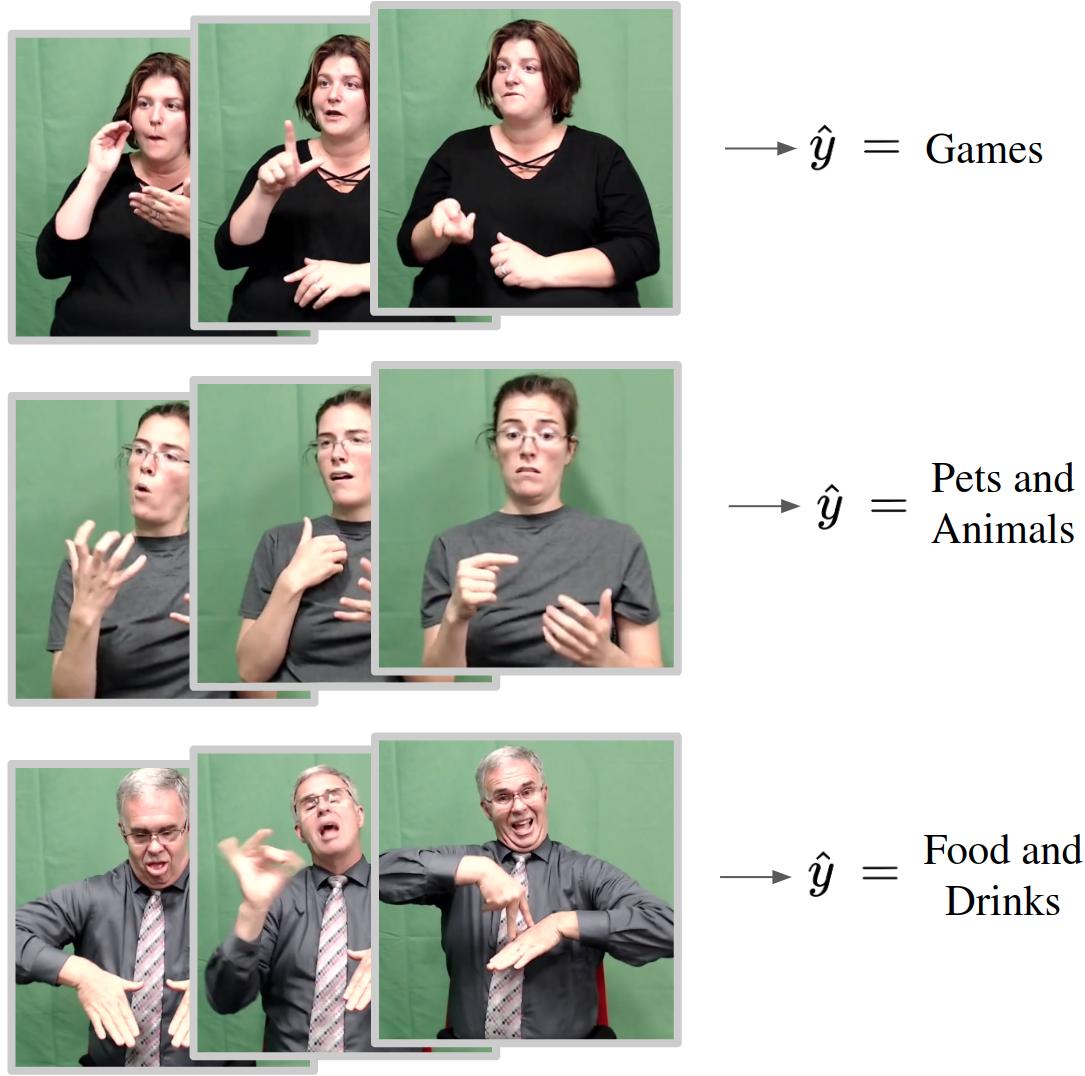}
    \end{center}
    \vspace{-0.35cm}
    \caption{Topic detection in sign language videos is the task of producing a label that describes the semantic content of a signer's discourse.}
    \label{fig:teaser}
    \vspace{-0.5cm}
\end{figure}

In this work, we propose the novel task of sign language topic detection, that is, classifying sign language video recordings into one of several categories, as depicted in Figure \ref{fig:teaser}. This task has been broadly explored for spoken languages \cite{TC_review20}, but not for sign languages.

We believe our work on topic detection in sign language videos could help in the design of more inclusive online experiences for the Deaf and hard-of-hearing. We tackle topic detection with three different neural architectures and three different kinds of video features, and evaluate their strengths and shortcomings through a set of experiments.

The contributions of this paper can be summarized as follows:

\begin{itemize}
  \item To the best of our knowledge, we provide the first study of sign language topic detection.
  \item We thoroughly measure the performance of three deep learning architectures (LSTM \cite{LSTM}, Transformer \cite{transformer} and PerceiverIO \cite{PerceiverIO}) in combination with three different video features that are commonly employed for sign language understanding (3D Cartesian body
poses, 3D angular body poses and I3D features).
    \item We make our code publicly available\footnote{ \href{https://github.com/imatge-upc/sign-topic}{https://github.com/imatge-upc/sign-topic}} implemented with Fairseq, a widely used toolkit for sequence modeling.
\end{itemize}

\section{Related Work}
\label{sec:related}

In this paper, we address the task of Sign Language (SL) topic detection, which we define as the task of producing a label that semantically describes the content of the discourse being signed, given a sequence of frames.

SL recognition is perhaps the closest task in the SL literature to that of topic detection. The aim in SL recognition is to tell which sign is being represented, given a short video of a signer producing either an isolated sign or a continuous sequence of signs \cite{Phoenix, SLT}.

The current state-of-the-art in SL recognition is characterized by complex modeling pipelines involving distillation \cite{SLRdistillation}, graph neural networks \cite{SkeletonSLR}, auxiliary losses \cite{SLR_VizAlign}, stochastic labeling \cite{SLR_stochastic}, or cross-modal alignment \cite{SLR_CrossModAlign}. In this work, however, we choose instead simpler pipelines that can act as robust baselines for future work on SL topic detection.

Outside the domain of SL, general video classification and action recognition is the most similar task to SL topic detection. Several methods have been proposed for generic video classification \cite{ViViT, VPN, UNIK, MultiViewT, CoCa, TokenLearner, VaTT}. Despite having obtained remarkable results, they are generally unsuitable for the task of topic detection we address here. Their computational requirements are often untenable, due to being designed for dealing with shorter videos of at most a few hundred frames, while SL videos may contain thousands of frames.

\begin{table*}[t]
    \centering
    \begin{tabular}{lcccc}
    \toprule
                    &\textbf{LSTM}      & \textbf{Transformer} & \textbf{Perceiver IO} \\
    \midrule
    Cartesian (OP)  & $32.85\pm5.21$    & $31.49\pm4.89$       & $30.78\pm2.95$ \\
    Cartesian (MP)  & $33.79\pm1.82$    & $32.64\pm2.41$       & $30.91\pm2.04$ \\
    \midrule
    Angular (OP)    & $32.64\pm3.80$    & $30.36\pm1.36$       & $30.22\pm1.10$ \\
    Angular (MP)    & $31.76\pm5.14$    & $32.41\pm3.83$       & $33.10\pm1.69$ \\
    \midrule
    I3D features    & $\bm{45.98\pm1.99}$    & $\bm{49.66\pm2.48}$  & $\bm{44.60\pm2.87}$ \\
    \midrule
    \midrule
    Transcriptions  & $70.31\pm2.97$    & $73.33\pm4.43$       & $71.13\pm2.83$ \\
    \bottomrule
    \vspace{-0.05cm}
    \end{tabular}
    \parbox{0.81\textwidth}{\caption{\label{tab:acc}Test accuracy (in \%) for each model and data type. We report the average and standard deviation over three runs. \textit{OP} stands for OpenPose and \textit{MP} for MediaPipe. Best results are in bold, transcription results are considered as a pseudo upper bound. Note that accuracy would be 25\% for majority class prediction.}}
    \vspace{-0.6cm}
\end{table*}

\section{Methodology}
\label{sec:methodology}

\subsection{Dataset}
\label{sec:dataset}

We base our experiments on the recent How2Sign dataset. How2Sign~\cite{Duarte_CVPR2021} is a large-scale collection of multimodal and multiview SL videos in American Sign Language (ASL) covering over 2500 instructional videos selected from the preexisting How2 dataset~\cite{How2}.

How2Sign consists of more than 80 hours of ASL videos, with sentence-level alignment for more than 35k sentences. It features a vocabulary of 16k English words that represent more than two thousand instructional videos from a broad range of categories. The dataset comes with a rich set of annotations including category labels, text annotations, as well automatically extracted 2D body poses for more than 6M frames.

\begin{figure}[H]
    \begin{center}
        \includegraphics[width=1\linewidth]{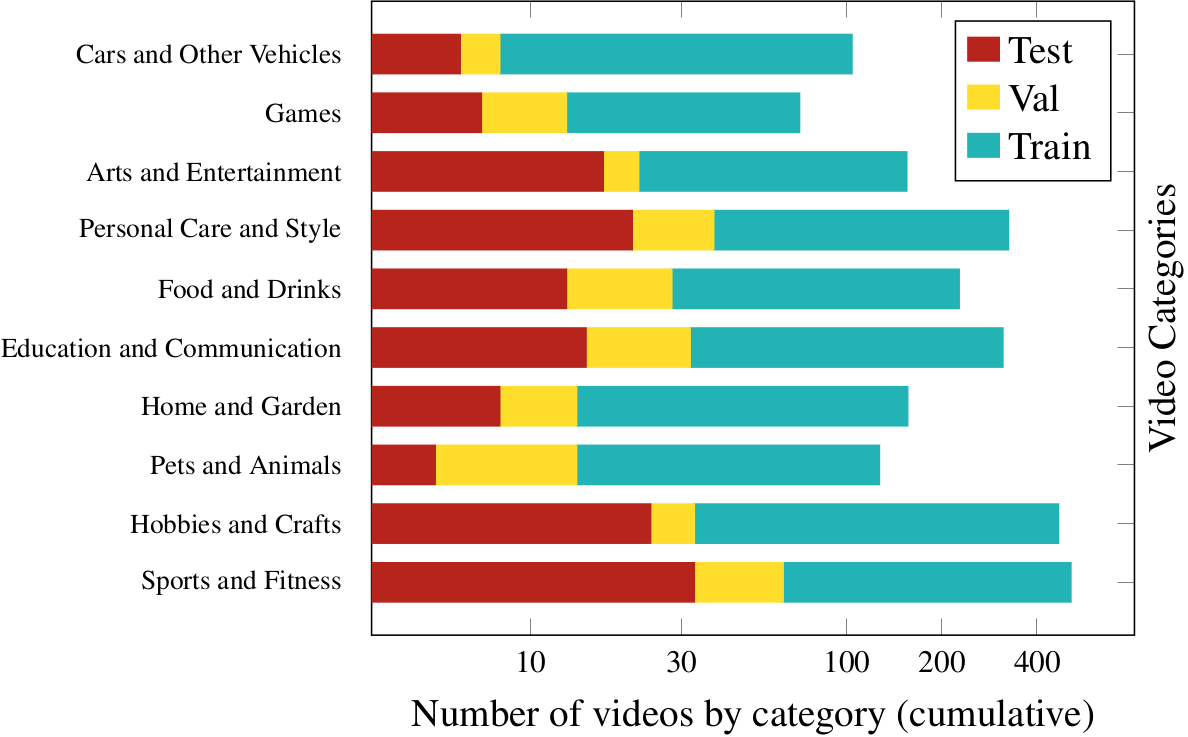}
    \end{center}
    \vspace{-0.27cm}
    \caption{Cumulative topic distribution in the How2Sign dataset (figure from \cite{Duarte_CVPR2021})}
    \label{fig:barchart}
    \vspace{-0.10cm}
\end{figure}

To the best of the authors' knowledge, How2Sign is currently the only SL dataset containing manually produced per-video category annotations semantically describing a video's content.
OpenASL \cite{OpenASL}, a recent large-scale ASL dataset, features 288 hours of SL video with speech transcriptions, but does not include category labels. Other datasets, such as \cite{Phoenix2, BOBSL} are restricted to a single topic or semantic domain. Others, like \cite{AUTSL}, just contain videos of isolated signs, rendering them unsuitable for SL video classification.  In this work, we leverage the topic annotations provided at video level (Fig. \ref{fig:barchart}). Each video is associated with one of 10 target labels describing its content, and our aim is to classify videos in their corresponding category.

\vspace{-0.05cm}

\subsection{Video Features}
\label{sec:features}

In our experiments, we train models with five different kinds of data: 3D poses represented by either Cartesian coordinates or joint angles, I3D features~\cite{I3D} and manually generated speech transcriptions, available from the How2 dataset~\cite{How2}. We include speech transcriptions to establish a pseudo-upper bound on performance, as we assume that a perfect sign language translation from the video sequences would match these speech transcriptions.

\textbf{3D Cartesian poses.}
Alongside videos, How2Sign also provides body keypoint annotations extracted with OpenPose \cite{OpenPose}. In addition, we also extract keypoints using MediaPipe \cite{MediaPipe}, resulting in two sets of body pose annotations.

\vspace{-0.05cm}
\begin{figure}[H]
    \begin{center}
        \includegraphics[width=0.95\linewidth]{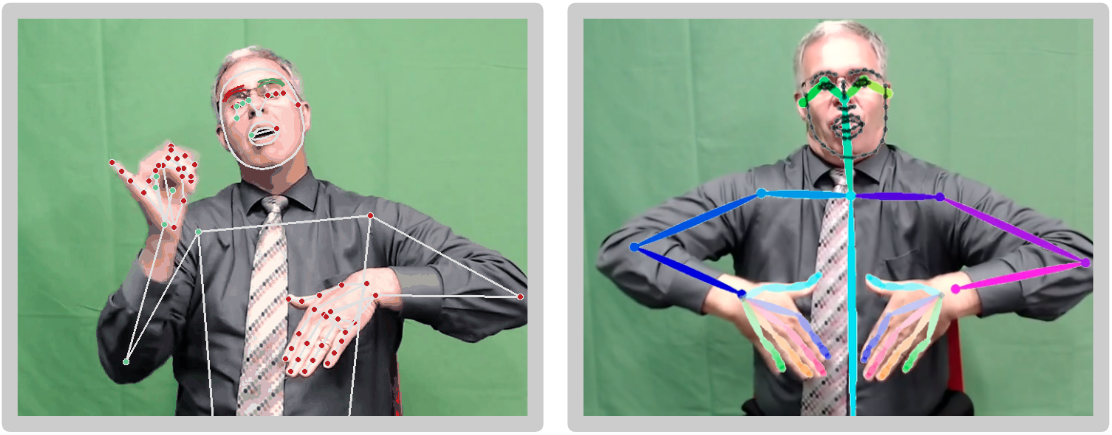}
    \end{center}
    \vspace{-0.3cm}
    \caption{We train models on body poses extracted with two different pose detectors: MediaPipe (left) and OpenPose (right).}
    \label{fig:mediapipe_openpose}
\end{figure}
\vspace{-0.25cm}

These keypoints provide a light-weight representation of the signer's hands and body that is invariant to the signers's and background's visual characteristics \cite{SLT_Korean}. Nevertheless, this representation is sensitive to occlusions and tends to present a significant amount of noise.
OpenPose and Mediapipe produce keypoints for hands, face and body, including arms and legs. We make use of the hands, upper body and arms. Since OpenPose does not produce 3D keypoints directly, we lift them to 3D as described in \cite{WordsGlosses}. Finally, we vectorize the pose for each frame into a vector $v_t = (x_1, y_1, z_1, ..., x_{50}, y_{50}, z_{50})$ of size $ 50 \times 3 = 150 $.

\textbf{3D angular poses.}
Although this 3D Cartesian representation allows handling occlusions and different camera angles much more effectively, it suffers from sensitivity to scale and length of the speaker's limbs. For this reason, we decided to follow~\cite{ng2021body2hands} and~\cite{Adam}, by converting the Cartesian coordinates to an angular representation~\cite{r6d_article}. In essence, this means that a vector $\theta_j$ associated with bone $j$ encodes the relative rotation of $j$ w.r.t its parent bones. For each frame, we vectorize the rotation vectors of each of the joints into a single array of size $48 \times 6 = 288$.

\textbf{I3D features fine-tuned for SLR.} We choose I3D features~\cite{I3D} to extract video representations directly from the RGB frames, motivated by their effectiveness in the sign recognition task~\cite{SL_retrieval, SLR_coArt, WLASL}. I3D features take into account not only visual cues, but also temporal information. As a result, they provide a dense and reliable source of visual cues as input to our models.

The original I3D network is trained on ImageNet~\cite{imagenet} and fine-tuned for action recognition with the Kinetics-400~\cite{Kinetics} dataset. As shown in~\cite{SL_retrieval}, further fine-tuning with sign language data is needed to properly model the temporal and spatial information present in them. We used the I3D features provided in~\cite{SL_retrieval}, which had been fine-tuned on the large-scale BBC-Oxford British Sign Language Dataset (BOBSL) for SLR.
We freeze the trained network to extract the visual features from the How2Sign videos and obtain the 1024-dimensional activation before the pooling layer of the I3D backbone.

\textbf{English transcriptions}
How2Sign provides English speech transcriptions in textual form for each of its videos. These transcriptions were manually produced and originate from the How2~\cite{How2} dataset, which How2Sign is based upon. English transcriptions were manually time-aligned at sentence-level with the How2Sign sign language videos. We embed the text into a $256$-dimensional trainable vector.

\subsection{Neural Architectures}
\label{sec:neural_arch}

We test three different architectures that stand behind some of the most notable successes in video analytics: the LSTM \cite{LSTM} with attention \cite{LSTMatt}, the Transformer \cite{transformer} and the PerceiverIO \cite{PerceiverIO}. These three architectures represent different trends for processing sequential inputs. The LSTM treats samples in a sequential manner, while the Transformer and the PerceiverIO process them in parallel via self-attention. PerceiverIO is specifically designed to handle extremely long input sequences, while the Transformer scales poorly with respect to input sequence length.

\textbf{LSTM with attention.}
LSTM is still one of the go-to architectures for dealing with sequential data. It processes an input video sequentially frame-by-frame, which allows it to scale linearly in terms of computational complexity with respect to the input length.

In order to boost the performance of the LSTM, we use a bidirectional configuration, and we add an attention mechanism over the hidden states, as described in \cite{LSTMatt}.

\begin{table*}[h]\footnotesize 
    \centering
    \begin{tabular}{clllllllll} 
    \toprule
    \multicolumn{1}{l}{} & \multicolumn{1}{c}{\rotatebox{90}{Params.}} & \multicolumn{1}{c}{\rotatebox{90}{FLOPs}} & \multicolumn{1}{c}{\rotatebox{90}{\begin{tabular}[c]{@{}c@{}}{Ratio}\\ {($10^{-4}$)}\end{tabular}}} &\multicolumn{1}{c}{\rotatebox{90}{Params.}} & \multicolumn{1}{c}{\rotatebox{90}{FLOPs}} & \multicolumn{1}{c}{\rotatebox{90}{\begin{tabular}[c]{@{}c@{}}{Ratio}\\ {($10^{-4}$)}\end{tabular}}} & \multicolumn{1}{c}{\rotatebox{90}{Params.}} & \multicolumn{1}{c}{\rotatebox{90}{FLOPs}} & \multicolumn{1}{c}{\rotatebox{90}{\begin{tabular}[c]{@{}c@{}}{Ratio}\\ {($10^{-4}$)}\end{tabular}}} \\
    \multicolumn{1}{l}{}  & \multicolumn{3}{c}{\textbf{LSTM}}      & \multicolumn{3}{c}{\textbf{Transformer}} & \multicolumn{3}{c}{\textbf{PerceiverIO}} \\
    \midrule
    \textbf{Cartesian (OP)}  & 9.6M  & 16.25B  & 5.91  & 4.5M   & 11.84B  & 3.80 & 1.0M  & 2.10B & 4.76 \\
    \textbf{Cartesian (MP)}  & 9.6M  & 16.25B  & 5.91  & 4.5M   & 11.84B  & 3.80 & 1.0M  & 2.10B & 4.76 \\
    \midrule
    \textbf{Angular (OP)}    & 10M   & 19.64B  & 5.19  & 5.2M   & 12.35B  & 4.21 & 1.1M  & 2.77B & 3.97 \\
    \textbf{Angular (MP)}    & 10M   & 19.64B  & 5.19  & 5.2M   & 12.35B  & 4.21 & 1.1M  & 2.77B & 3.97 \\
    \midrule
    \textbf{I3D features}    & 13M   & 31.41B  & 4.18  & 8.9M   & 15.21B  & 5.85 & 3.1M  & 4.76B & 6.51 \\
    \midrule
    \midrule
    \textbf{Transcriptions}  & 12M   & 18.86B  & 6.38  & 4.4M   & 12.24B  & 3.09 & 1.5M  & 2.79B & 5.38 \\
    \bottomrule
    \end{tabular}
    \caption{\label{tab:params_flops} Number of parameters and FLOPs for each model and data type. The ratio measures the amount of parameters per FLOP. \textit{OP} stands for OpenPose and \textit{MP} for MediaPipe.}
    \vspace{-0.35cm}
\end{table*}

\textbf{Transformer.}
Since its appearance \cite{transformer}, the Transformer has dominated the NLP landscape and has also recently become prominent on several image processing tasks \cite{nlp_survey}. The core component of the Transformer is the self-attention module which performs a comparison of each of the input tokens against the rest. One advantage of the Transformer over the LSTM, is that the Transformer allows processing input tokens in parallel in a non-sequential fashion, thus reducing training time. However, Transformer incurs in a quadratic computational cost with respect to the input length.

\textbf{PerceiverIO.}
A recent trend in the machine learning literature is to design deep learning architectures that overcome the quadratic cost of the self-attention mechanism. One line of work has focused on projecting the inputs to a lower dimensional latent space. PerceiverIO \cite{PerceiverIO} leverages a cross-attention module at the beginning of the architecture which maps an input array of length $T$ to a latent array of length $N$, with $N \ll T$.

\section{Experiments}
\label{sec:exp}

A suite of models are trained across several architectures and feature types, with the aim of introducing strong baseline models with different characteristics for the task of sign language topic detection. For each pair of architecture and data type, we perform a grid search in order to select the most adequate hyperparameters.

We train all of our models on a single GeForce RTX 3090 GPU. We run training until validation accuracy stops decreasing and use early stopping on validation accuracy to select the best checkpoint. As optimizer, we utilize Adam \cite{AdamOptim} with a learning rate scheduler having a decrease factor of 0.5 per 8 epochs of non-decreasing validation loss. We leave the learning rate as a hyperparameter to be determined for each model.

We use the SentencePiece \cite{sentence_piece} tokenizer with a dictionary size of 8000 for speech transcriptions and 1470 for spotted signs, and input embeddings of size 256 for all models using one of these two input types. We implement our models and training pipelines on the Fairseq library \cite{fairseq}, which runs on PyTorch and is designed to perform translation, summarization and other spoken language tasks.

The baselines for the topic detection task are presented in Table~\ref{tab:acc}. Addressing the task with the manually generated speech transcriptions is the best performing approach. All models trained on speech transcriptions obtain a test accuracy of over 70\%, with Transformer obtaining the highest score. These values establish an upper-bound for this task and dataset.

Among the video features, the pretrained I3D features yield the highest test accuracy across all architectures, with the Transformer at the top with almost 50\% accuracy on the 10-class classification problem.
The accuracy obtained with body poses are consistently poorer than than those with I3D, but the results are still well above random predictions.
Notice though that the I3D features had been fine-tuned for the SL recognition task over the BOBSL dataset, while body poses are extracted off-the-shelf and not explicitly optimized for sign language video understanding.

All body pose features yield similar results, with no clear indication that angular poses might be more suitable than Cartesian ones, or that either one of the pose extractors (OpenPose and MediaPipe) is more adequate than the other. Nevertheless, representing the signers' bodies with keypoints tends to give poorer results. We consider that this gap in performance is related to the limitations of the body pose estimators against the fast motion and self-occlusions of the hands common in sign language videos. Moreover, none of the studied architectures are specifically designed for processing body pose inputs, which can be naturally described in the form of a graph, rather than an array of values, as we do in this work.

As seen in Table \ref{tab:params_flops}, PerceiverIO has the lowest amount of parameters per floating point operation (FLOP), and the LSTM has the highest amount of FLOPs. This is mainly due to the fact that the LSTM processes a whole sequence of length $T$, while the Transformer processes a downsampled sequence of length smaller than $T$, and the PerceiverIO reduces the initial sequence of length $T$ to a much shorter latent sequence.

\section{Conclusions}
\label{sec:conclusions}

In this work, we present the task of sign language topic detection for the first time in the literature. We provide baseline models for topic detection in sign language videos, which will contribute to the design of more inclusive experiences for the Deaf and hard-of-hearing.

We have found that I3D features fine-tuned for sign language recognition lead to better performance for the task, while body poses lag far behind in terms of accuracy, but are also more generic, privacy-preserving and lighter to analyze. 

Future work should aim at improving the quality of visual features, by developing pose estimation systems paired with other neural architectures with more appropriate inductive biases, and by leveraging model pre-training on adjacent SL tasks and datasets.

{\small
\bibliographystyle{ieee_fullname}
\bibliography{egbib}

\begin{thebibliography}{10}\itemsep=-1pt

\bibitem{VaTT}
Hassan Akbari, Liangzhe Yuan, Rui Qian, Wei-Hong Chuang, Shih-Fu Chang, Yin
  Cui, and Boqing Gong.
\newblock Vatt: Transformers for multimodal self-supervised learning from raw
  video, audio and text.
\newblock {\em Advances in Neural Information Processing Systems}, 34, 2021.

\bibitem{SLR_coArt}
Samuel Albanie, G{\"u}l Varol, Liliane Momeni, Triantafyllos Afouras, Joon~Son
  Chung, Neil Fox, and Andrew Zisserman.
\newblock Bsl-1k: Scaling up co-articulated sign language recognition using
  mouthing cues.
\newblock In {\em European Conference on Computer Vision}, pages 35--53.
  Springer, 2020.

\bibitem{BOBSL}
Samuel Albanie, G{\"u}l Varol, Liliane Momeni, Hannah Bull, Triantafyllos
  Afouras, Himel Chowdhury, Neil Fox, Bencie Woll, Rob Cooper, Andrew
  McParland, and Andrew Zisserman.
\newblock {BOBSL}: {BBC}-{O}xford {B}ritish {S}ign {L}anguage {D}ataset.
\newblock 2021.

\bibitem{ViViT}
Anurag Arnab, Mostafa Dehghani, Georg Heigold, Chen Sun, Mario Lu{\v{c}}i{\'c},
  and Cordelia Schmid.
\newblock Vivit: A video vision transformer.
\newblock In {\em Proceedings of the IEEE/CVF International Conference on
  Computer Vision}, pages 6836--6846, 2021.

\bibitem{Phoenix2}
Necati~Cihan Camgoz, Simon Hadfield, Oscar Koller, Hermann Ney, and Richard
  Bowden.
\newblock Neural sign language translation.
\newblock In {\em 2018 IEEE/CVF Conference on Computer Vision and Pattern
  Recognition}, pages 7784--7793, 2018.

\bibitem{SLT_MultiChannel}
Necati~Cihan Camgoz, Oscar Koller, Simon Hadfield, and Richard Bowden.
\newblock Multi-channel transformers for multi-articulatory sign language
  translation.
\newblock In {\em European Conference on Computer Vision}, pages 301--319.
  Springer, 2020.

\bibitem{OpenPose}
Z. {Cao}, G. {Hidalgo Martinez}, T. {Simon}, S. {Wei}, and Y.~A. {Sheikh}.
\newblock Openpose: Realtime multi-person 2d pose estimation using part
  affinity fields.
\newblock {\em IEEE Transactions on Pattern Analysis and Machine Intelligence},
  2019.

\bibitem{I3D}
Joao Carreira and Andrew Zisserman.
\newblock Quo vadis, action recognition? a new model and the kinetics dataset.
\newblock In {\em proceedings of the IEEE Conference on Computer Vision and
  Pattern Recognition}, pages 6299--6308, 2017.

\bibitem{Kinetics}
Joao Carreira and Andrew Zisserman.
\newblock Quo vadis, action recognition? a new model and the kinetics dataset.
\newblock In {\em Proceedings of the IEEE Conference on Computer Vision and
  Pattern Recognition (CVPR)}, July 2017.

\bibitem{SLT}
Necati Cihan~Camgöz, Oscar Koller, Simon Hadfield, and Richard Bowden.
\newblock Sign language transformers: Joint end-to-end sign language
  recognition and translation.
\newblock In {\em 2020 IEEE/CVF Conference on Computer Vision and Pattern
  Recognition (CVPR)}, pages 10020--10030, 2020.

\bibitem{VPN}
Srijan Das, Rui Dai, Di Yang, and Francois Bremond.
\newblock Vpn++: Rethinking video-pose embeddings for understanding activities
  of daily living.
\newblock {\em IEEE Transactions on Pattern Analysis and Machine Intelligence},
  2021.

\bibitem{RGB_SLR}
Mathieu De~Coster, Mieke Van~Herreweghe, and Joni Dambre.
\newblock Isolated sign recognition from rgb video using pose flow and
  self-attention.
\newblock In {\em 2021 IEEE/CVF Conference on Computer Vision and Pattern
  Recognition Workshops (CVPRW)}, pages 3436--3445, 2021.

\bibitem{imagenet}
Jia Deng, Wei Dong, Richard Socher, Li-Jia Li, Kai Li, and Li Fei-Fei.
\newblock Imagenet: A large-scale hierarchical image database.
\newblock In {\em 2009 IEEE conference on computer vision and pattern
  recognition}, pages 248--255. Ieee, 2009.

\bibitem{SL_retrieval}
Amanda Duarte, Samuel Albanie, Xavier Gir{\'o}-i Nieto, and G{\"u}l Varol.
\newblock Sign language video retrieval with free-form textual queries.
\newblock In {\em Conference on Computer Vision and Pattern Recognition
  (CVPR)}, 2022.

\bibitem{Duarte_CVPR2021}
Amanda Duarte, Shruti Palaskar, Lucas Ventura, Deepti Ghadiyaram, Kenneth
  DeHaan, Florian Metze, Jordi Torres, and Xavier Giro-i Nieto.
\newblock {How2Sign: A Large-scale Multimodal Dataset for Continuous American
  Sign Language}.
\newblock In {\em Conference on Computer Vision and Pattern Recognition
  (CVPR)}, 2021.

\bibitem{SLRdistillation}
Aiming Hao, Yuecong Min, and Xilin Chen.
\newblock Self-mutual distillation learning for continuous sign language
  recognition.
\newblock In {\em 2021 IEEE/CVF International Conference on Computer Vision
  (ICCV)}, pages 11283--11292, 2021.

\bibitem{LSTM}
Sepp Hochreiter and Jürgen Schmidhuber.
\newblock Long short-term memory.
\newblock {\em Neural Computation}, 9(8):1735--1780, 1997.

\bibitem{PerceiverIO}
Andrew Jaegle, Sebastian Borgeaud, Jean{-}Baptiste Alayrac, Carl Doersch,
  Catalin Ionescu, David Ding, Skanda Koppula, Daniel Zoran, Andrew Brock, Evan
  Shelhamer, Olivier~J. H{\'{e}}naff, Matthew~M. Botvinick, Andrew Zisserman,
  Oriol Vinyals, and Jo{\~{a}}o Carreira.
\newblock Perceiver {IO:} {A} general architecture for structured inputs {\&}
  outputs.
\newblock In {\em International Conference on Learning Representations (ICLR)},
  2022.

\bibitem{SkeletonSLR}
Songyao Jiang, Bin Sun, Lichen Wang, Yue Bai, Kunpeng Li, and Yun Fu.
\newblock Skeleton aware multi-modal sign language recognition.
\newblock In {\em Challenge on Large Scale Signer Independent Isolated Sign
  Language Recognition (CVPR)}, 2021.

\bibitem{AdamOptim}
Diederik~P. Kingma and Jimmy Ba.
\newblock Adam: A method for stochastic optimization.
\newblock In {\em International Conference on Learning Representations (ICLR)},
  2015.

\bibitem{SLT_Korean}
Sang-Ki Ko, Chang~Jo Kim, Hyedong Jung, and Choongsang Cho.
\newblock Neural sign language translation based on human keypoint estimation.
\newblock {\em Applied Sciences}, 9(13):2683, 2019.

\bibitem{Phoenix}
Oscar Koller, Jens Forster, and Hermann Ney.
\newblock Continuous sign language recognition: Towards large vocabulary
  statistical recognition systems handling multiple signers.
\newblock {\em Computer Vision and Image Understanding}, 141:108--125, Dec.
  2015.

\bibitem{sentence_piece}
Taku Kudo and John Richardson.
\newblock {S}entence{P}iece: A simple and language independent subword
  tokenizer and detokenizer for neural text processing.
\newblock In {\em Proceedings of the 2018 Conference on Empirical Methods in
  Natural Language Processing: System Demonstrations}, pages 66--71, Brussels,
  Belgium, Nov. 2018. Association for Computational Linguistics.

\bibitem{WLASL}
Dongxu Li, Cristian Rodriguez, Xin Yu, and Hongdong Li.
\newblock Word-level deep sign language recognition from video: A new
  large-scale dataset and methods comparison.
\newblock In {\em Proceedings of the IEEE/CVF winter conference on applications
  of computer vision}, pages 1459--1469, 2020.

\bibitem{MediaPipe}
Camillo Lugaresi, Jiuqiang Tang, Hadon Nash, Chris McClanahan, Esha Uboweja,
  Michael Hays, Fan Zhang, Chuo-Ling Chang, Ming~Guang Yong, Juhyun Lee, et~al.
\newblock Mediapipe: A framework for building perception pipelines.
\newblock {\em arXiv preprint arXiv:1906.08172}, 2019.

\bibitem{SLR_VizAlign}
Yuecong Min, Aiming Hao, Xiujuan Chai, and Xilin Chen.
\newblock Visual alignment constraint for continuous sign language recognition.
\newblock In {\em Proceedings of the IEEE/CVF International Conference on
  Computer Vision}, pages 11542--11551, 2021.

\bibitem{TC_review20}
Shervin Minaee, Nal Kalchbrenner, Erik Cambria, Narjes Nikzad, Meysam
  Chenaghlu, and Jianfeng Gao.
\newblock Deep learning--based text classification: a comprehensive review.
\newblock {\em ACM Computing Surveys (CSUR)}, 54(3):1--40, 2021.

\bibitem{ng2021body2hands}
Evonne Ng, Shiry Ginosar, Trevor Darrell, and Hanbyul Joo.
\newblock Body2hands: Learning to infer 3d hands from conversational gesture
  body dynamics.
\newblock {\em Proceedings of the IEEE/CVF Conference on Computer Vision and
  Pattern Recognition}, pages 11865--11874, 2021.

\bibitem{SLR_stochastic}
Zhe Niu and Brian Mak.
\newblock Stochastic fine-grained labeling of multi-state sign glosses for
  continuous sign language recognition.
\newblock In Andrea Vedaldi, Horst Bischof, Thomas Brox, and Jan-Michael Frahm,
  editors, {\em Computer Vision -- ECCV 2020}, pages 172--186, Cham, 2020.
  Springer International Publishing.

\bibitem{fairseq}
Myle Ott, Sergey Edunov, Alexei Baevski, Angela Fan, Sam Gross, Nathan Ng,
  David Grangier, and Michael Auli.
\newblock fairseq: A fast, extensible toolkit for sequence modeling.
\newblock In {\em Proceedings of the 2019 Conference of the North {A}merican
  Chapter of the Association for Computational Linguistics (Demonstrations)},
  pages 48--53, Minneapolis, Minnesota, June 2019. Association for
  Computational Linguistics.

\bibitem{SLR_CrossModAlign}
Ilias Papastratis, Kosmas Dimitropoulos, Dimitrios Konstantinidis, and Petros
  Daras.
\newblock Continuous sign language recognition through cross-modal alignment of
  video and text embeddings in a joint-latent space.
\newblock {\em IEEE Access}, 8:91170--91180, 2020.

\bibitem{TokenLearner}
Michael~S. Ryoo, A.~J. Piergiovanni, Anurag Arnab, Mostafa Dehghani, and Anelia
  Angelova.
\newblock Tokenlearner: What can 8 learned tokens do for images and videos?
\newblock In {\em Proceedings of the Conference on Neural Information
  Processing Systems (NeurIPS)}, 2021.

\bibitem{How2}
Ramon Sanabria, Ozan Caglayan, Shruti Palaskar, Desmond Elliott, Lo{\"\i}c
  Barrault, Lucia Specia, and Florian Metze.
\newblock How2: a large-scale dataset for multimodal language understanding.
\newblock {\em arXiv preprint arXiv:1811.00347}, 2018.

\bibitem{SLT_EvSignNow}
Ben Saunders, Necati~Cihan Camgoz, and Richard Bowden.
\newblock Everybody sign now: Translating spoken language to photo realistic
  sign language video.
\newblock {\em arXiv preprint arXiv:2011.09846}, 2020.

\bibitem{SLP_ProgTransf}
Ben Saunders, Necati~Cihan Camgoz, and Richard Bowden.
\newblock Continuous 3d multi-channel sign language production via progressive
  transformers and mixture density networks.
\newblock {\em International journal of computer vision}, 129(7):2113--2135,
  2021.

\bibitem{SLP_Mixed}
Ben Saunders, Necati~Cihan Camgoz, and Richard Bowden.
\newblock Mixed signals: Sign language production via a mixture of motion
  primitives.
\newblock In {\em Proceedings of the IEEE/CVF International Conference on
  Computer Vision}, pages 1919--1929, 2021.

\bibitem{SLP_GraphSAtt}
Ben Saunders, Necati~Cihan Camgoz, and Richard Bowden.
\newblock Skeletal graph self-attention: Embedding a skeleton inductive bias
  into sign language production.
\newblock {\em arXiv preprint arXiv:2112.05277}, 2021.

\bibitem{SLP_Photo}
Ben Saunders, Necati~Cihan Camgoz, and Richard Bowden.
\newblock Signing at scale: Learning to co-articulate signs for large-scale
  photo-realistic sign language production.
\newblock {\em arXiv preprint arXiv:2203.15354}, 2022.

\bibitem{OpenASL}
Bowen Shi, Diane Brentari, Greg Shakhnarovich, and Karen Livescu.
\newblock Open-domain sign language translation learned from online video.
\newblock {\em arXiv preprint arXiv:2205.12870}, 2022.

\bibitem{AUTSL}
Ozge~Mercanoglu Sincan and Hacer~Yalim Keles.
\newblock Autsl: A large scale multi-modal turkish sign language dataset and
  baseline methods.
\newblock {\em IEEE Access}, 8:181340--181355, 2020.

\bibitem{transformer}
Ashish Vaswani, Noam Shazeer, Niki Parmar, Jakob Uszkoreit, Llion Jones,
  Aidan~N Gomez, {\L}ukasz Kaiser, and Illia Polosukhin.
\newblock Attention is all you need.
\newblock {\em Advances in neural information processing systems}, 30, 2017.

\bibitem{Adam}
Donglai Xiang, Hanbyul Joo, and Yaser Sheikh.
\newblock Monocular total capture: Posing face, body, and hands in the wild.
\newblock In {\em Proceedings of the IEEE/CVF conference on computer vision and
  pattern recognition}, pages 10965--10974, 2019.

\bibitem{MultiViewT}
Shen Yan, Xuehan Xiong, Anurag Arnab, Zhichao Lu, Mi Zhang, Chen Sun, and
  Cordelia Schmid.
\newblock Multiview transformers for video recognition.
\newblock In {\em Conference on Computer Vision and Pattern Recognition
  (CVPR)}, 2022.

\bibitem{UNIK}
Di Yang, Yaohui Wang, Antitza Dantcheva, Lorenzo Garattoni, Gianpiero
  Francesca, and Fran{\c{c}}ois Br{\'{e}}mond.
\newblock {UNIK:} {A} unified framework for real-world skeleton-based action
  recognition.
\newblock In {\em British Machine Vision Conference (BMVC)}, 2021.

\bibitem{CoCa}
Jiahui Yu, Zirui Wang, Vijay Vasudevan, Legg Yeung, Mojtaba Seyedhosseini, and
  Yonghui Wu.
\newblock Coca: Contrastive captioners are image-text foundation models.
\newblock {\em arXiv preprint arXiv:2205.01917}, 2022.

\bibitem{WordsGlosses}
Jan Zelinka and Jakub Kanis.
\newblock Neural sign language synthesis: Words are our glosses.
\newblock In {\em 2020 IEEE Winter Conference on Applications of Computer
  Vision (WACV)}, pages 3384--3392, 2020.

\bibitem{nlp_survey}
Ming Zhou, Nan Duan, Shujie Liu, and Heung-Yeung Shum.
\newblock Progress in neural nlp: Modeling, learning, and reasoning.
\newblock {\em Engineering}, 6(3):275--290, 2020.

\bibitem{LSTMatt}
Peng Zhou, Wei Shi, Jun Tian, Zhenyu Qi, Bingchen Li, Hongwei Hao, and Bo Xu.
\newblock Attention-based bidirectional long short-term memory networks for
  relation classification.
\newblock In {\em Proceedings of the 54th Annual Meeting of the Association for
  Computational Linguistics (Volume 2: Short Papers)}, pages 207--212, Berlin,
  Germany, Aug. 2016. Association for Computational Linguistics.

\bibitem{r6d_article}
Yi Zhou, Connelly Barnes, Jingwan Lu, Jimei Yang, and Hao Li.
\newblock On the continuity of rotation representations in neural networks.
\newblock In {\em 2019 IEEE/CVF Conference on Computer Vision and Pattern
  Recognition (CVPR)}, pages 5738--5746, 2019.

\end{thebibliography}
}

\end{document}